\documentclass{article}
\usepackage[nonatbib,final]{main}

\usepackage[utf8]{inputenc} 
\usepackage[T1]{fontenc}
\usepackage[numbers]{natbib}
\usepackage{url}    
\usepackage{hyperref}
\usepackage{xcolor}
\definecolor{darkblue}{rgb}{0, 0.40, 0.75}
\hypersetup{colorlinks=true, citecolor=darkblue, linkcolor=darkblue, urlcolor=darkblue}

\usepackage[utf8]{inputenc} 
\usepackage[T1]{fontenc}    
\usepackage{hyperref}      
\usepackage{url}       
\usepackage{booktabs}    
\usepackage{amsfonts}     
\usepackage{nicefrac}     
\usepackage{microtype}    
\usepackage{xcolor}       
\usepackage{amsmath}
\usepackage{graphicx}
\usepackage{subcaption}
\usepackage{amsmath}
\usepackage{wrapfig}
\usepackage{adjustbox} 
\usepackage{multirow}
\usepackage{paralist}
\usepackage{CJKutf8}
\usepackage{subcaption}
\usepackage{colortbl}
\usepackage{tgpagella} 
\usepackage{mathpazo}  
\usepackage{XCharter} 
\usepackage[textsize=tiny]{todonotes}
\usepackage{arydshln}
\usepackage[most]{tcolorbox}
\usepackage{enumitem}

\definecolor{newgreen}{rgb}{0.7, 0.9, 0.7}
\definecolor{newblue}{rgb}{0.85, 0.85, 0.9}
\definecolor{hidden-red}{RGB}{205, 44, 36}
\definecolor{hidden-blue}{RGB}{194,232,247}
\definecolor{hidden-orange}{RGB}{243,202,120}
\definecolor{hidden-green}{RGB}{34,139,34}
\definecolor{hidden-pink}{RGB}{255,245,247}
\definecolor{hidden-black}{RGB}{20,68,106}
\definecolor{purple}{RGB}{144,153,196}
\definecolor{yellow}{RGB}{255,228,123}
\definecolor{hidden-yellow}{RGB}{255,248,203}
\definecolor{tkcolor}{RGB}{224,223,255}
\definecolor{darkblue}{rgb}{0, 0.40, 0.75}
\newcommand{\modelname}{\texttt{IF-Track}}
\definecolor{tkcolor}{RGB}{224,223,255}
\definecolor{shallowred}{RGB}{242,204,208}
\tcbset{
  takeawaybox/.style={
    width=\linewidth,
    top=8pt,
    bottom=4pt,
    colback=blue!6!white,
    colframe=black,
    colbacktitle=black,
    enhanced,
    center,
    attach boxed title to top left={yshift=-0.1in,xshift=0.15in},
    boxed title style={boxrule=0pt,colframe=white,},
  }
}

\newtcolorbox{TakeawayBox}[2][]{takeawaybox,title=#2,#1}

\title{\fontsize{15pt}{15pt}\selectfont The Universal Landscape of Human Reasoning}

\author{
Qiguang Chen$^{1}$\thanks{Equal Contribution} \quad Jinhao Liu$^{1}$\footnotemark[1] \quad Libo Qin$^{2}$\thanks{Corresponding Author} \quad Yimeng Zhang$^{3}$ \quad Yihao Liang$^{4}$ \\
\textbf{Shangxu Ren$^{1}$ \quad Chengyu Luan$^{1}$ \quad Dengyun Peng$^{1}$ \quad Hanjing Li$^{1}$ \quad Jiannan Guan$^{1}$} \\
\textbf{Zheng Yan$^{1}$ \quad Jiaqi Wang$^{5}$ \quad Mengkang Hu$^{6}$ \quad Yantao Du$^{7}$ \quad Zhi Chen$^{7}$  } \\\vspace{5pt}
\textbf{Xie Chen$^{8}$ \quad Wanxiang Che$^{1}$\footnotemark[2]} \\
$^{1}$ Harbin Institute of Technology \quad $^{2}$ Central South University \\
$^{3}$ University of Illinois Urbana-Champaign \quad
$^{4}$ Princeton University \\
$^{5}$ The Chinese University of Hong Kong \quad
$^{6}$ The University of Hong Kong \\
$^{7}$ ByteDance Seed (China) \quad $^{8}$ Shanghai Jiao Tong University \\
\texttt{\{qgchen,jhliu,car\}@ir.hit.edu.cn},  \texttt{qinlibo@hit.edu.cn} \\
}

\begin{document}
	
\maketitle
	
\begin{abstract}
Understanding how information is dynamically accumulated and transformed in human reasoning has long challenged cognitive psychology, philosophy, and artificial intelligence. Existing accounts, from classical logic to probabilistic models, illuminate aspects of output or individual modelling, but do not offer a unified, quantitative description of general human reasoning dynamics. To solve this, we introduce Information Flow Tracking (\modelname{}), that uses large language models (LLMs) as probabilistic encoder to quantify information entropy and gain at each reasoning step. Through fine-grained analyses across diverse tasks, our method is the \textit{first successfully models the universal landscape of human reasoning behaviors} within a single metric space. We show that \modelname{} captures essential reasoning features, identifies systematic error patterns, and characterizes individual differences. Applied to discussion of advanced psychological theory, we first reconcile single- versus dual-process theories in \modelname{} and discover the alignment of artificial and human cognition and how LLMs reshaping human reasoning process. This approach establishes a quantitative bridge between theory and measurement, offering mechanistic insights into the architecture of reasoning.

\vspace{5pt}
\end{abstract}
	
\section{Introduction}

Human reasoning modelling has long been central to cognitive psychology, philosophy, and artificial intelligence, addressing how reasoning processes are structured~\citep{storring1908experimentelle,wagman1997cognitive,wagman2002problem,holyoak2005cambridge,furbach2017bridging,zhuang2023through,chen2025towards}. Early approaches grounded in classical logic, such as propositional and deductive reasoning, modeled cognition through fixed rules and formal structures~\citep{storring1908experimentelle,stenning2012human}. With the rise of probabilistic paradigms, attention shifted to models incorporating heuristic and uncertainty-based mechanisms, exemplified by Bayesian reasoning frameworks~\citep{oaksford2007bayesian,chater2010bayesian,johnson2015logic,stenseke2024computational}.
In contrast, the theory of mental models posits that reasoning involves constructing and manipulating possible situations rather than adhering to static rules~\citep{johnson2006we}. Recent works refine this view, emphasizing reasoning as the evaluation of possibilities and necessities beyond formal logic~\citep{johnson2024models}. More recently, meta-learning approaches highlight reasoning as an adaptive process, where cognitive abilities evolve through interaction with the environment, providing a dynamic account of individual human cognition~\citep{Binz_Dasgupta_Jagadish_Botvinick_Wang_Schulz_2024,Maier2025,Ramadan2025,Bruckner2025}.

From a neuroscience perspective, reasoning engages specific brain regions. Evidence indicates that human reasoning activates the prefrontal and parietal cortices, with theta oscillations in the prefrontal cortex linked to thought processes~\citep{amin2014brain}. When reasoning output is correct, beta activity in EEG recordings increases significantly~\citep{salto2023electrical}, providing biological support for reasoning models and underscoring the central role of neural dynamics in reasoning.
Nevertheless, most current approaches focus on endpoint performance and isolated measures, lacking a unified quantitative framework to track general reasoning trajectories continuously. This limitation constrains mechanistic, process-level insight and obscures temporally resolved features, including the errors, types, and individual features of dynamic reasoning landscapes.

\begin{figure*}[t]
    \centering
    \includegraphics[width=\linewidth]{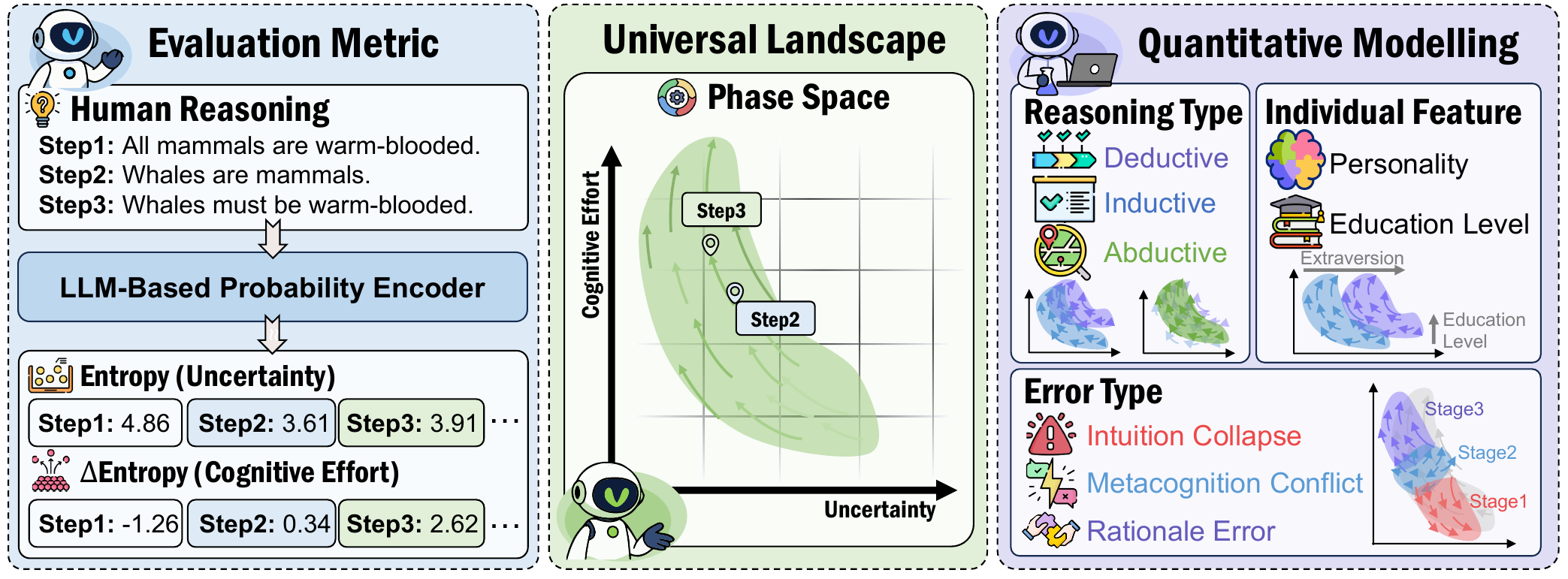}
    \caption{
    \textbf{Theoretical Framework and Modelling Applications of \modelname{}.}\\
    \textit{
    \textbf{a.}  Computation of reasoning metrics.
    An LLM-based probability encoder estimates the conditional probability of each reasoning step, yielding two variables: \emph{uncertainty} (information entropy, \(u_t\)) and \emph{cognitive effort} (information gain, \(e_t\)).
    \textbf{b.} Theoretical foundation: Information phase space.
    Reasoning can be modeled as a trajectory within a two-dimensional \emph{information phase space} \((u_t, e_t)\), formulated as a Hamiltonian system that models the reasoning landscape. This framework depicts the transition from high uncertainty and low effort (Step 2) to low uncertainty and high effort (Step 3).
    \textbf{c.} Applications under the framework.
    Built on this framework, \modelname{} models reasoning across \emph{Reasoning Type} (e.g. Deductive and Inductive Reasoning), \emph{Individual Feature} (e.g., personality, professional background), and
    \emph{Error Type} (e.g., intuition collapse, metacognitive conflict, rationale error).
    \vspace{-6pt}}
    }
    \label{fig:method}
\end{figure*}

To address this gap, as illustrated in Fig.~\ref{fig:method}\textit{a}, we introduce a framework that first treats large language models (LLMs) as probabilistic encoders to quantify the landscape of human reasoning by tracking, step-by-step along reasoning trajectories, information entropy (uncertainty), and the relevance development value (cognitive effort). Furthermore, as shown in Fig.~\ref{fig:method}\textit{b}, to our knowledge, \textbf{Information Flow Tracking (\modelname{}) framework delivers the first universal landscape of human reasoning} across diverse tasks through a unified information phase space. This formulation produces reproducible, cross-task signatures of cognitive landscape, enabling precise comparisons of reasoning strategies and uncovering previously inaccessible patterns of information flow. These insights offer substantial implications for understanding human cognition in societal contexts and advancing methodologies in cognitive and social sciences.

Further, as shown in Fig.~\ref{fig:method}\textit{c}, we present a unified modelling framework that captures distinct reasoning patterns and models stepwise errors, showing how errors in intermediate states shape subsequent stages. Beyond feature modelling, it also models stable individual signatures, revealing how personality and educational background influence information processing and reasoning paths.
These offer theoretical guidance for dissecting human reasoning behaviors and practical strategies for refining reasoning in large models.
We apply this framework to key debates in cognitive psychology, including single- versus dual-process theories, which diverge locally yet converge globally. Concurrently, we show how LLMs reshape human reasoning, yielding insights on the evolution between human cognition and AI.

In summary, our contributions are as follows:
\begin{itemize}[leftmargin=2ex,topsep=0pt]
    \item \textbf{Providing Universal Landscape:} To the best of our knowledge, we are the first to quantitatively model universal human reasoning landscape, providing a new framework for quantitative analysis of reasoning behavior.
    \item \textbf{Effectively Modelling Reasoning Features:} We effectively capture and model key features of human reasoning processes, representing inductive and deductive reasoning in two distinct modes and integrating abductive reasoning through their combination.
    \item \textbf{Successfully Analyzing Individual Differences:} We quantify behavioral variation across individuals differing in personality and professional background, providing fresh insights into how such factors shape information processing and path selection.
    \item \textbf{Quantitative Application of Psychological Theory:} We apply this framework to discussions of psychological theories, such as single- versus dual-process models of reasoning, which differ locally but align globally. In parallel, we contrast how LLMs reshape human reasoning, yielding insights for aligning human cognition with AI.
\end{itemize}

\section{Theory Model}
\label{sec:theory}
To model the landscape of human reasoning, this section introduces the theoretical foundations of the \textbf{Information Flow Tracking (\modelname{})} framework, which model information changes governed by Hamiltonian dynamics. This approach demonstrates stable information flow during reasoning, analogous to physical systems. See Sec.~\ref{sec:model} for a detailed description.\vspace{-8pt}

\paragraph{Hamiltonian Dynamics} 
Generally, the evolution of a physical system can be elegantly formulated within \textbf{Hamiltonian dynamics}~\cite{goldstein2002classical,landau1976mechanics,arnold1989mathematical}. 
At any given time step \( t \), the system’s state is characterized by a pair of conjugate variables, the \textit{generalized coordinate} \( q_t \) and the \textit{generalized momentum} \( p_t \). 
Their temporal evolution follows Hamilton’s canonical equations:
\begin{equation}
\dot{q}_t = \frac{\partial H}{\partial p_t}, \quad 
\dot{p}_t = -\frac{\partial H}{\partial q_t},
\end{equation}
where \( H(q_t, p_t) \) denotes the Hamiltonian function, typically representing the total energy of the system. 
This formulation describes a \textit{conservative flow} in which total energy is preserved and the trajectory evolves deterministically in the phase space. 
Beyond physical systems, in information phase space, uncertainty and cognitive effort can be modeled as conjugate variables reflecting informational energy and mental motion~\cite{jaynes1957information,liang2005information,friston2010freeenergy}.
\vspace{-8pt}

\paragraph{Information Flow Tracking}
Further, we define \textbf{Information Flow Tracking (\modelname{})} within an \textbf{information phase space} to describe reasoning as a continuous cognitive flow based on Hamiltonian dynamics. As shown in Fig.~\ref{fig:method}\textit{a}, 
given each reasoning step \( t \), the cognitive state is represented by a pair \((u_t, e_t)\), where \( u_t \) denotes the \textit{uncertainty} (quantified by information entropy) and \( e_t \) denotes the \textit{cognitive effort} (measured by the change in entropy, i.e., information gain between steps).
The reasoning dynamics thus form a trajectory \( \mathbf{X}_t = (u_t, e_t) \), which evolves according to an underlying information flow field \( \dot{\mathbf{X}}_t = \mathbf{V}(u_t, e_t) \).

As shown Fig.~\ref{fig:method}\textit{b}, each reasoning process can be viewed as a transition between two cognitive states (\textit{Step1 $\rightarrow$ Step2}) in this two-dimensional space: 
it typically begins in a high-uncertainty, low-effort region (\(u_1\) high, \(e_1\) low), reflecting intuitive exploration; 
and moves toward a low-uncertainty, high-effort region (\(u_2 < u_1,\, e_2 > e_1\)), representing deliberate analysis. 
This transition mirrors a \textit{conservative evolution} in which total ``informational energy'' remains constant but redistributes between uncertainty and effort, an information-theoretic analogue to energy–momentum exchange in Hamiltonian dynamics.
\vspace{-8pt}

\paragraph{Information Flow Tracking meets Hamiltonian Dynamics.}
According to \textbf{Liouville’s theorem}, Hamiltonian flow in phase space is divergence-free:
\begin{equation}
\nabla \!\cdot \! \vec{V} =
\frac{\partial \dot{q}_t}{\partial q_t} +
\frac{\partial \dot{p}_t}{\partial p_t} = 0,
\end{equation}
implying that the total phase-space volume remains conserved during evolution~\cite{arnold1989mathematical}.
We extend this conservation principle to human reasoning in \modelname{}: 
Within the \textit{information phase space} defined by uncertainty \(u_t\) and cognitive effort \(e_t\), 
the reasoning information flow field \( \vec{V}(u_t, e_t) = (\dot{u}_t, \dot{e}_t) \) satisfies:
\begin{equation}
\nabla \!\cdot \! \vec{V}(u_t,e_t) = 0,
\end{equation}
indicating that human reasoning maintains \textbf{a conserved structure in its information dynamics (viz, information flow field is phase space)}. 
Empirically, it manifests in the smooth, divergence-free trajectories, where reasoning evolves continuously from intuitive to analytical states without loss of information. See more theoretical analyses in the Sec.~\ref{sec:model}.

\begin{figure*}[t]
    \centering
    \includegraphics[width=\linewidth]{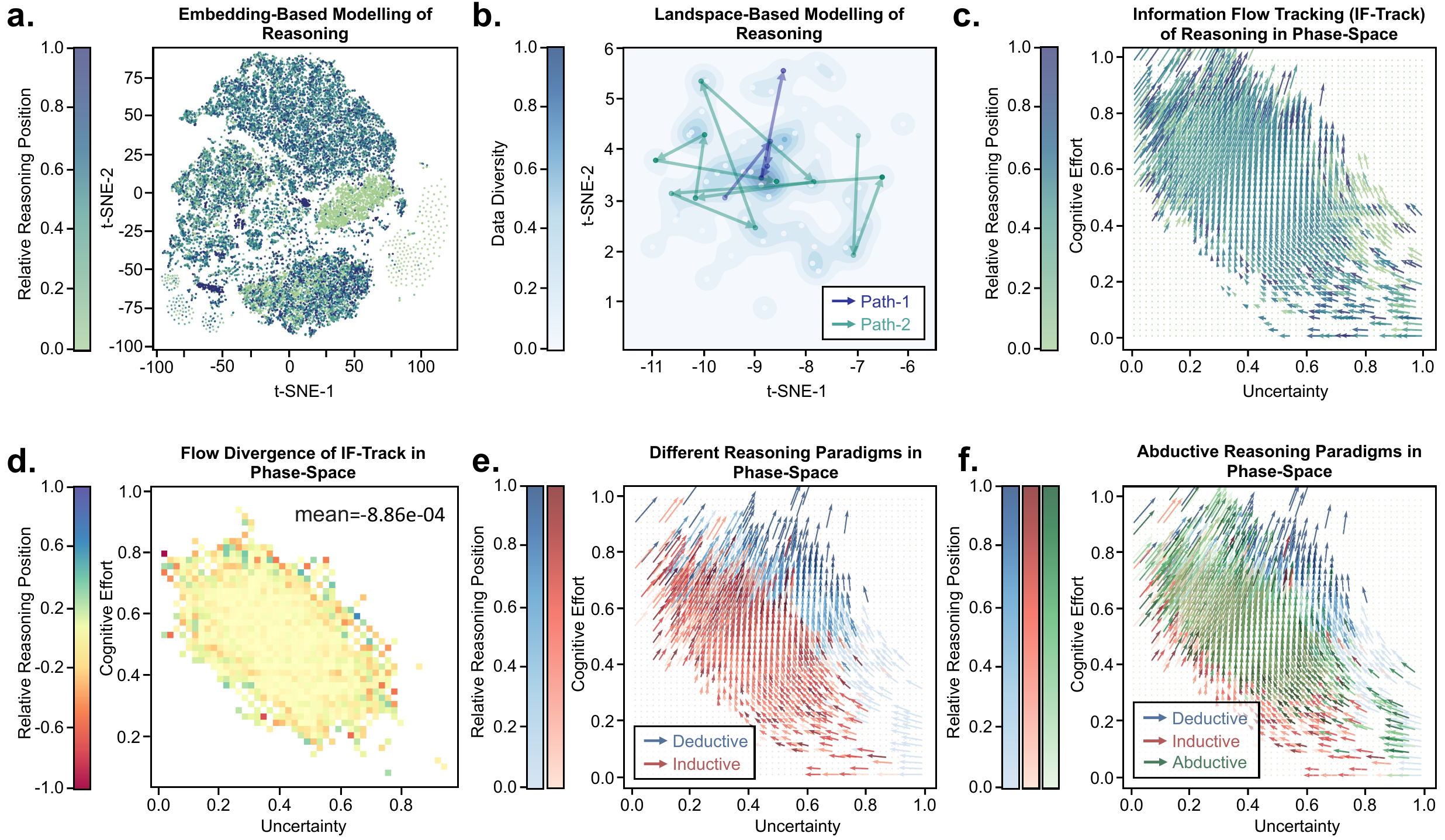}
    \caption{
        \textbf{Comparison of static representations of reasoning trajectories and relevant reasoning paradigms modelling.}\\
        \textit{\textbf{a.} t-SNE projection of step embeddings, showing chaotic representations across different human reasoning processes.
        \textbf{b.} Visualization of the landscape of thought, showing clustered but temporally unordered patterns across different human reasoning processes.
        \textbf{c.} Reasoning trajectories in the entropy–gain phase space reveal a globally consistent flow, where arrows represent the direction of information flow, illustrating the dynamic evolution of reasoning states from high uncertainty toward stable cognitive states.
        \textbf{d.} Empirical validation of the information phase space. The pseudocolor map shows the local divergence \(\nabla\cdot\vec{V}\) of the reasoning flow in the \((u,e)\) space. Most regions exhibit near-zero divergence (uniform color), indicating approximate volume preservation and supporting a quasi-Hamiltonian structure of human reasoning dynamics.
        \textbf{e.} Different reasoning paradigms (deductive vs. inductive reasoning) in phase space.
        \textbf{f.} Abductive reasoning in phase space, lying between deductive and inductive and showing a hybrid pattern.\vspace{-2mm}}
    }
    \label{fig:verify}
\end{figure*}

\section{Results}

\subsection{Universal Human Reasoning Landscape Modelling}
In this section, we validate the core theoretical claims of the \modelname{} framework introduced in Sec.~\ref{sec:theory} by empirically testing whether it can successfully quantify and track human reasoning dynamics as an approximately incompressible information flow in phase space.\vspace{-8pt}

\paragraph{Current static modelling methods can not unify modelling human reasoning landscape.}
Most existing reasoning modelling methods emphasize static semantic distributions or output-only results, offering static snapshots rather than a dynamic process of reasoning~\citep{fintz2022deep,Yang2025DynamicEarlyExit,CDMRNet2025}.
As depicted in Fig.~\ref{fig:verify}\textit{a}, embedding the reasoning steps enables static visualizations of reasoning representations. 
Projecting these embeddings with t-SNE~\citep{maaten2008visualizing} produces clustered patterns but removes temporal order, obscuring how the thought process unfolds~\citep{SplusTSNE2024,VisualTemporal2024}.
The more advanced ``landscape of thought'' approach~\citep{zhou2025landscape} renders these static representations as smooth surfaces for multiple-choice tasks; however, for general reasoning, as shown in Fig.~\ref{fig:verify}\textit{b}, trajectories (green and purple) vary widely across problems, which undermines both the consistency and the interpretability of the modelling. Hence, neither their sequential order nor shared structure can be captured by those two static modelling methods, because the visualizations remain irregular and overlapping.\vspace{-8pt}

\begin{wrapfigure}{r}{0.5\textwidth}
\centering
\includegraphics[width=0.46\textwidth]{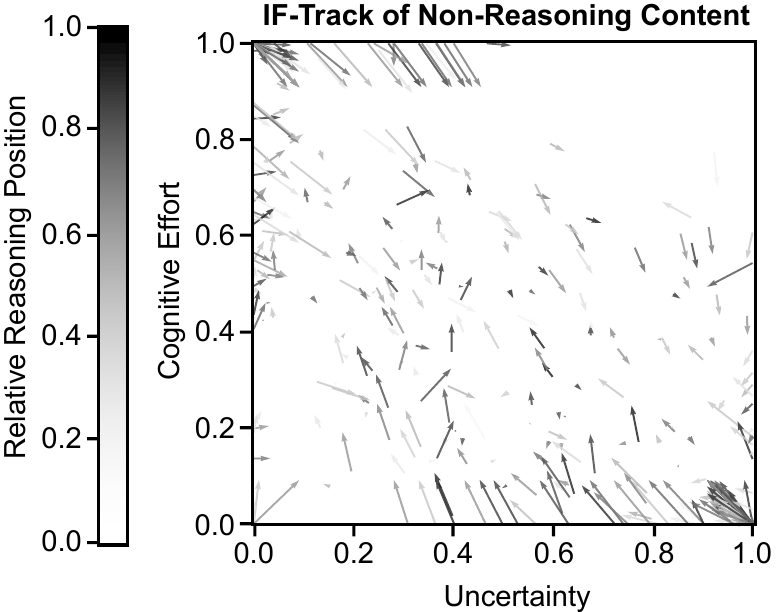}
\caption{
\textbf{\modelname{} of non-reasoning data.}\\
\textit{
The trajectory distribution of non-reasoning data (\(n=1054\)) shows no consistent directional flow in the uncertainty–effort phase space, 
indicating the absence of structured reasoning dynamics.\vspace{-8pt}
}}
\label{fig:reasoning_verification}
\end{wrapfigure}

\paragraph{Human reasoning process can be successfully quantified and tracked by \modelname{}.}
By mapping reasoning steps to the normalized phase space, as shown in Fig.~\ref{fig:verify}\textit{c}, \modelname{} establishes an "information phase space" where the indicated arrow represents a consistent flow direction. This approach maintains coherent flow that enables both progression and interpretability. In contrast, non-reasoning scenarios (Fig.~\ref{fig:reasoning_verification} in Methods) exhibit disordered dynamics. Thus, \modelname{} quantifies reasoning trajectories as structured flow fields with consistent information paths. Notably, these flows show distinct dynamics: uncertainty decreases as intermediate conclusions accumulate, rebounding slightly at the final integrative step due to synthesis-induced doubt or further exploration; meanwhile, cognitive effort rises steadily. Overall, \modelname{} integrates reasoning into a unified dynamic framework that captures the temporal directionality and structural essence of human thought.\vspace{-8pt}

\paragraph{Human reasoning as an approximately incompressible information flow satisfying Liouville's equation in phase space.}
To test whether the inferred cognitive dynamics are quasi-Hamiltonian and consistent with Liouville's equation, we computed the local divergence \( \nabla \cdot \vec{V} \) along reasoning trajectories and visualized it as a pseudocolor map in Fig.~\ref{fig:verify}\textit{d}. Extended regions of nearly uniform color indicate near-zero divergence (as yellow color in the Figure; mean<1e-3), consistent with an approximately volume-preserving flow in phase space. Small deviations appear primarily at the boundaries, suggesting weak dissipative effects attributable to noise or boundary interactions. These observations support the theoretical soundness of our phase-space modelling of reasoning trajectories. Overall, the evidence is consistent with a quasi-Hamiltonian description in which uncertainty and cognitive effort act as effective conjugate variables that trade off while approximately conserving phase-space volume.

\subsection{Reasoning Classical Attribution Modelling}

To evaluate whether \modelname{} can model classical attribution, we analyze both distinct reasoning types and common reasoning errors. Our framework shows capability in both respects:
(1) it distinguishes classical reasoning types via trajectory patterns (Sec.~\ref{sec:trajectory_patterns});
(2) it identifies and classifies reasoning errors as deviations from typical trajectories (Sec.~\ref{sec:error_classification}).

\subsubsection{\modelname{} Distinguishes Classical Reasoning Types via Trajectory Patterns.}
\label{sec:trajectory_patterns}
Human reasoning is traditionally classified into three fundamental types in cognitive psychology~\citep{peirce1931collected,harman1965inference,johnson1983formal,rips1994cognitive,douven2017abduction}:
(1) \textbf{Deductive reasoning} derives conclusions that necessarily follow from explicit premises, guaranteeing truth when the premises are true.
(2) \textbf{Inductive reasoning} generalizes from specific observations to broader principles, yielding probabilistic rather than certain conclusions.
(3) \textbf{Abductive reasoning} infers the most plausible explanations for incomplete or surprising evidence, supporting hypothesis generation and diagnostic inference.
While these categories have long guided qualitative studies, whether \modelname{} can model dynamic distinctions quantitatively is still an important question.\vspace{-8pt}

\paragraph{Deductive and Inductive reasoning exhibit similar global patterns but distinct local dynamics.}
As shown in Fig.~\ref{fig:verify}\textit{e}, both deductive and inductive reasoning follow a similar global pattern: uncertainty drops sharply at the outset, stabilizes midway, and slightly rebounds near the end, while cognitive effort rises steadily throughout.
Yet their local dynamics diverge.
Deductive reasoning starts with higher cognitive effort and rapid uncertainty reduction, consistent with its rule-based, top-down character.
Inductive reasoning, by contrast, begins with lower effort and slower uncertainty reduction, reflecting exploratory pattern discovery.
This difference supports the cognitive view that deduction and induction share a common structure but differ in their early-stage processing dynamics.\vspace{-8pt}

\paragraph{Abductive Reasoning works with a hybrid dynamic pattern.}  
As illustrated in Fig.~\ref{fig:verify}\textit{f}, abductive reasoning occupies an intermediate position between deduction and induction on both uncertainty and effort.
Its global trend mirrors the other two: uncertainty declines and then slightly rebounds, while cognitive effort accumulates.
Early steps show moderate uncertainty and low effort, consistent with tentative hypothesis formation.
Subsequent steps alternate between surges in effort and shifts in uncertainty, reflecting iterative hypothesis testing and refinement.  
This hybrid trajectory presents abduction as a synthesis of exploratory inference (as in induction) and confirmatory reasoning (as in deduction), consistent with classical accounts of abductive cognition~\citep{Peirce1931,Josephson1994-JOSAIC,JohnsonLaird1999,Magnani2004-MAGMAM}.

\begin{figure*}[t]
  \centering
  \includegraphics[width=\linewidth]{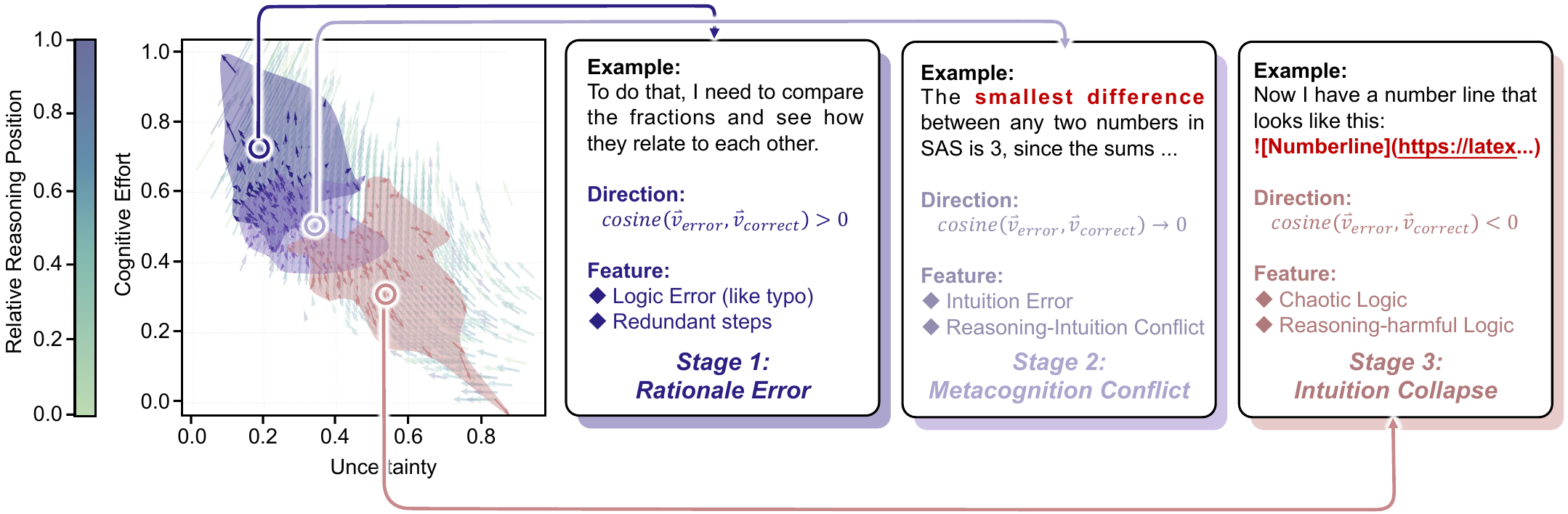}
  \caption{Three categories of reasoning errors identified by \modelname{}, positioned in the uncertainty–effort  phase space. ($n_{total}$=9,991, $n_{error}=374$). These errors were clustered into three stages by directions: \textit{Stage~1:~Intuition~Collapse} (Ratio~87.3\%), 
\textit{Stage~2:~Metacognition~Conflict} (Ratio~73.7\%), and 
\textit{Stage~3:~Rationale~Error} (Ratio~90.4\%)
\vspace{-6pt}}
  \label{fig:error_flow}
\end{figure*}

\subsubsection{\modelname{} effectively identifies reasoning errors based on trajectory deviations.}
\label{sec:error_classification}
Beyond classifying reasoning categories, a robust modelling approach must evaluate step-level correctness. To assess this capability, we analyze about 9,991 human-annotated reasoning steps that include 372 annotated erroneous steps across multiple categories. As shown in Fig.~\ref{fig:error_flow}, these errors cluster into three stages consistent with Pennycook’s three-stage theory of reasoning errors~\cite{PENNYCOOK2023131}. Each stage maps to a distinct region defined by spatial position and directional dynamics of the reasoning trajectory, enabling \modelname{} to identify error types from trajectory signatures. Specifically, these stages show the following features:
\begin{itemize}[leftmargin=2ex,topsep=0pt]
    \item \textbf{Stage 1: Intuition Collapse} is located in the lower-right corner of the phase space and marking the start of the reasoning flow with high uncertainty and low cognitive effort. Trajectories are impulsive and disorganized, often reversing direction (\( \cos(\vec{v}_{\mathrm{error}}, \vec{v}_{\mathrm{correct}}) < 0 \)), indicating motion opposite to the correct trajectory. These errors arise from faulty or unfounded intuition, where reasoning collapses before monitoring or deliberation.
    \item \textbf{Stage 2: Metacognition Conflict} is positioned in the central band of the phase space with moderate uncertainty and effort. It captures reasoning that appears coherent but rests on flawed assumptions. The direction cosine (\( \cos(\vec{v}_{\mathrm{error}}, \vec{v}_{\mathrm{correct}}) \approx 0 \)) indicates lateral divergence from the correct flow rather than reversal. Such errors reflect conflict-monitoring failures, where inconsistencies or contradictions go unnoticed during mid-stage reasoning.
    \item \textbf{Stage 3: Rationale Error} can be found in the upper-left area with low uncertainty but high cognitive effort. Reasoning remains aligned with the correct trajectory (\( \cos(\vec{v}_{\mathrm{error}}, \vec{v}_{\mathrm{correct}}) > 0 \)) yet suffers from inefficient or minor error processing, such as redundancy, over-explanation, or arithmetic slips, after the correct structure is established.
\end{itemize}
Together, these categories show that reasoning errors are systematically distributed across the uncertainty–effort phase space. This alignment provides empirical support that \modelname{} not only detects reasoning failures but also recapitulates the cognitive progression described in human reasoning theories.

\begin{figure*}[t]
    \centering
    \includegraphics[width=\linewidth]{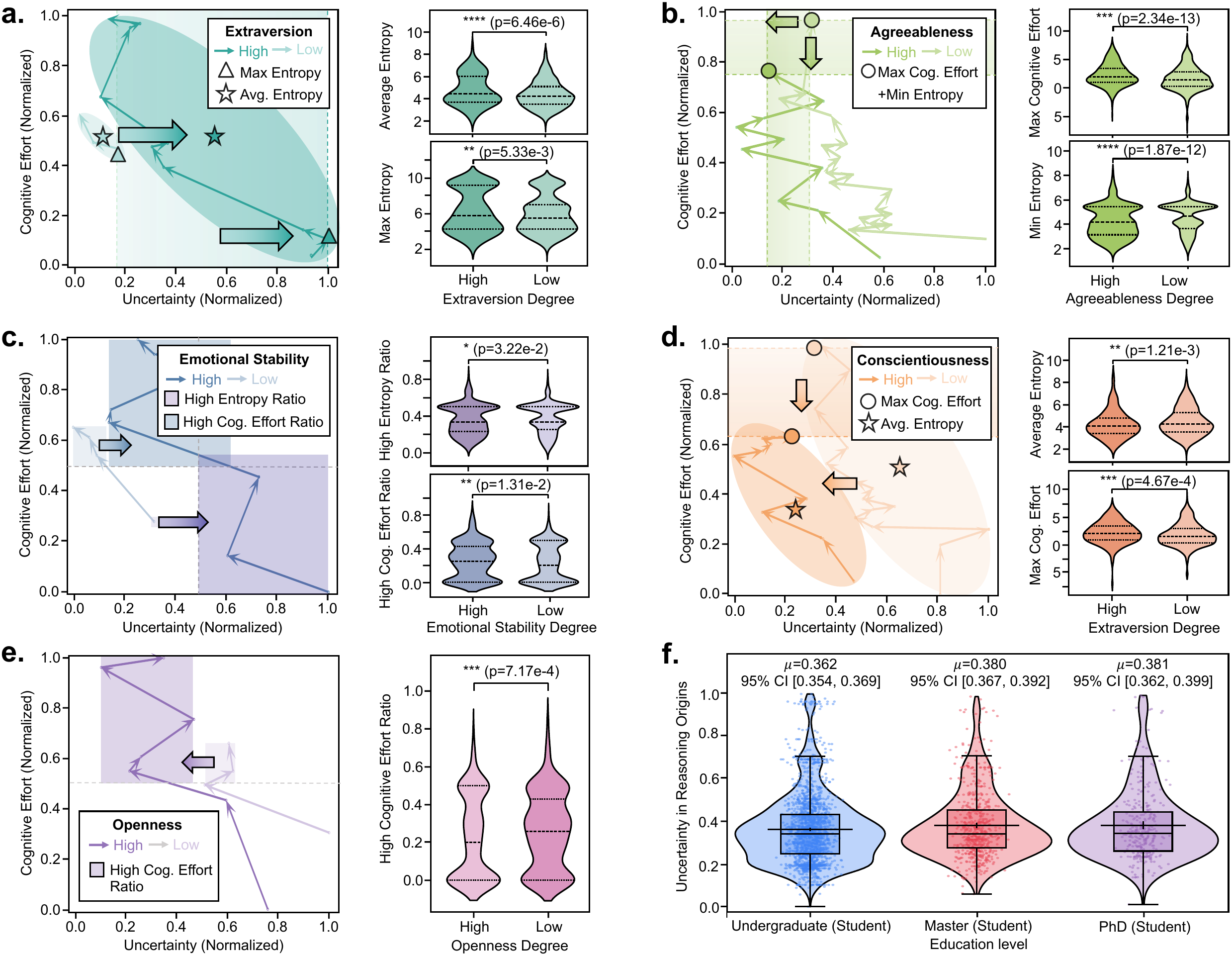}
    \caption{
        \textbf{Personality-related modulation of reasoning trajectories and cognitive–informational dynamics.}\\
        \textit{
        \textbf{a.} Extraversion. Individuals with higher extraversion exhibit greater mean and maximal entropy, indicating higher tolerance for uncertainty and broader exploratory reasoning. 
        \textbf{b.} Conscientiousness. The high-conscientiousness group shows lower average entropy but higher maximal effort, consistent with disciplined and goal-oriented reasoning. 
        \textbf{c.} Emotional Stability. Individuals with higher emotional stability maintain a more balanced entropy–effort profile, reflected in higher ratios of high-entropy and high-effort states. 
        \textbf{d.} Openness. Higher openness corresponds to a greater proportion of high-effort reasoning phases, suggesting deeper cognitive engagement and flexible information integration. 
        \textbf{e.} Agreeableness. High-agreeableness participants show higher maximal cognitive effort and lower minimal entropy, suggesting more sustained and focused reasoning. 
        \textbf{f.} Education Level. Across undergraduate, master, and PhD groups, higher educational attainment correlates with slightly higher uncertainty at reasoning origins, indicating broader hypothesis search spaces in early-stage reasoning. 
        \textbf{Total } \(n=3{,}215\).
        \vspace{-6pt}
        }
    }

    \label{fig:human}
\end{figure*}

\subsection{Effective Individual Characteristics Modelling}

Beyond characterizing general reasoning patterns, our framework captures \emph{individual variability} in reasoning behavior.
We collected 6,452 reasoning trajectories from participants worldwide, spanning deductive, inductive, and abductive tasks.
The dataset integrates demographic and psychological attributes, with emphasis on personality traits and education level, to test whether \modelname{} quantitatively reveals individual characteristics.

\subsubsection{Personality Traits Modelling}
Personality traits modulate how individuals engage with \textbf{Uncertainty} and \textbf{Cognitive Effort} during reasoning.
In this study, we adopt the Big Five Personality Traits as a comprehensive framework to characterize individual differences.
This model captures five relatively independent dimensions, \textit{Extraversion}, \textit{Conscientiousness}, \textit{Agreeableness}, \textit{Emotional Stability}, and \textit{Openness}, that jointly describe variations in affective tendencies, motivation, and cognitive processing.
\vspace{-8pt}

\paragraph{Extraversion: Preference for High-Uncertainty Exploration.}
Arousal theory~\citep{eysenck1991dimensions} posits that extraverts have lower baseline cortical arousal and therefore seek stimulation. As shown in Fig.~\ref{fig:human}\textit{a}, individuals high in Extraversion exhibit higher average Uncertainty (\(p=6.46\times 10^{-6}\)) and maximal Uncertainty (\(p=5.33\times 10^{-3}\)), consistent with a preference for ambiguous or unpredictable states and with evidence that extraversion and positive mood sustain persistence under ambiguity~\citep{hirsh2008predicting}.\vspace{-8pt}

\paragraph{Agreeableness: Seeking Efficient and Certain Trajectories.}
In Fig.~\ref{fig:human}\textit{b}, higher Agreeableness is associated with greater maximal Cognitive Effort (\(p=2.34\times 10^{-13}\)) and lower maximal Uncertainty (\(p=1.87\times 10^{-12}\)). Participants higher in Agreeableness tend to initiate reasoning in stable, low-Uncertainty states and then increase Cognitive Effort along structured trajectories, consistent with certainty-seeking, consensus-oriented, and conflict-averse processing characteristic of this trait~\citep{byulibrary2008five}.\vspace{-8pt}

\paragraph{Emotional Stability: High Uncertainty Tolerance and Efficient Reasoning.}
As shown in Fig.~\ref{fig:human}\textit{c}, higher Emotional Stability is associated with higher proportions of high-uncertainty and high-cognitive-effort states (\(p<0.05\)). This pattern is consistent with evidence that individuals with high Emotional Stability tolerate uncertainty and ambiguity while maintaining coherent and efficient reasoning~\citep{ford2001stress}.\vspace{-8pt}

\paragraph{Conscientiousness: Structured, Goal-Oriented Reasoning.}
In Fig.~\ref{fig:human}\textit{d}, participants high in conscientiousness show lower average uncertainty and higher peak cognitive effort (\(p<0.01\)). 
These individuals follow more structured reasoning paths, exhibiting reduced variability and uncertainty while concentrating peak cognitive effort at critical junctures~\citep{costa1992neo}.\vspace{-8pt}

\paragraph{Openness: Greater Exploratory Engagement.}
Fig.~\ref{fig:human}\textit{e} indicates that higher Openness scores are associated with a greater proportion of high cognitive-effort states (\(p=7.17\times 10^{-4}\)), while Uncertainty remains stable. This pattern suggests that individuals higher in Openness explore a broader range of reasoning paths without increasing Uncertainty~\citep{MCCRAE1997825,DEYOUNG2002533,deyoung2015openness}.

\subsubsection{Educational Level Modelling}
\paragraph{Educational attainment appears to shape not only knowledge but also the initial conditions of reasoning.} 
Fig.~\ref{fig:human}\textit{f} compares the uncertainty of the first reasoning step across undergraduate, master's, and PhD participants. 
Higher education levels correspond to greater \textbf{initial uncertainty} (\(\mu_{\text{PhD}}=0.381 > \mu_{\text{Undergrad}}=0.362\)), and the 95\% confidence intervals overlap, indicating a consistent improvement. 
This pattern suggests that advanced academic training may encourage reasoning from broader hypothesis spaces, with reduced reliance on prior knowledge, greater self-directed exploration, and a higher tolerance for ambiguity at early stages~\cite{cattell1987intelligence,Stanovich_West_2000}.

\subsection{Application \modelname{} for Adavanced Psychological Theory Discussion}

Based on previous anlaysis, we robustly conceptualise human reasoning as dynamic information flow in \modelname{} within an information phase space. Building on this, we apply \modelname{} to psychological theory, including dual-process accounts and human–LLM alignment.

\subsubsection{Single- vs. Dual-Process Theories Debates}

\begin{figure*}[t]
    \centering
    \includegraphics[width=0.98\linewidth]{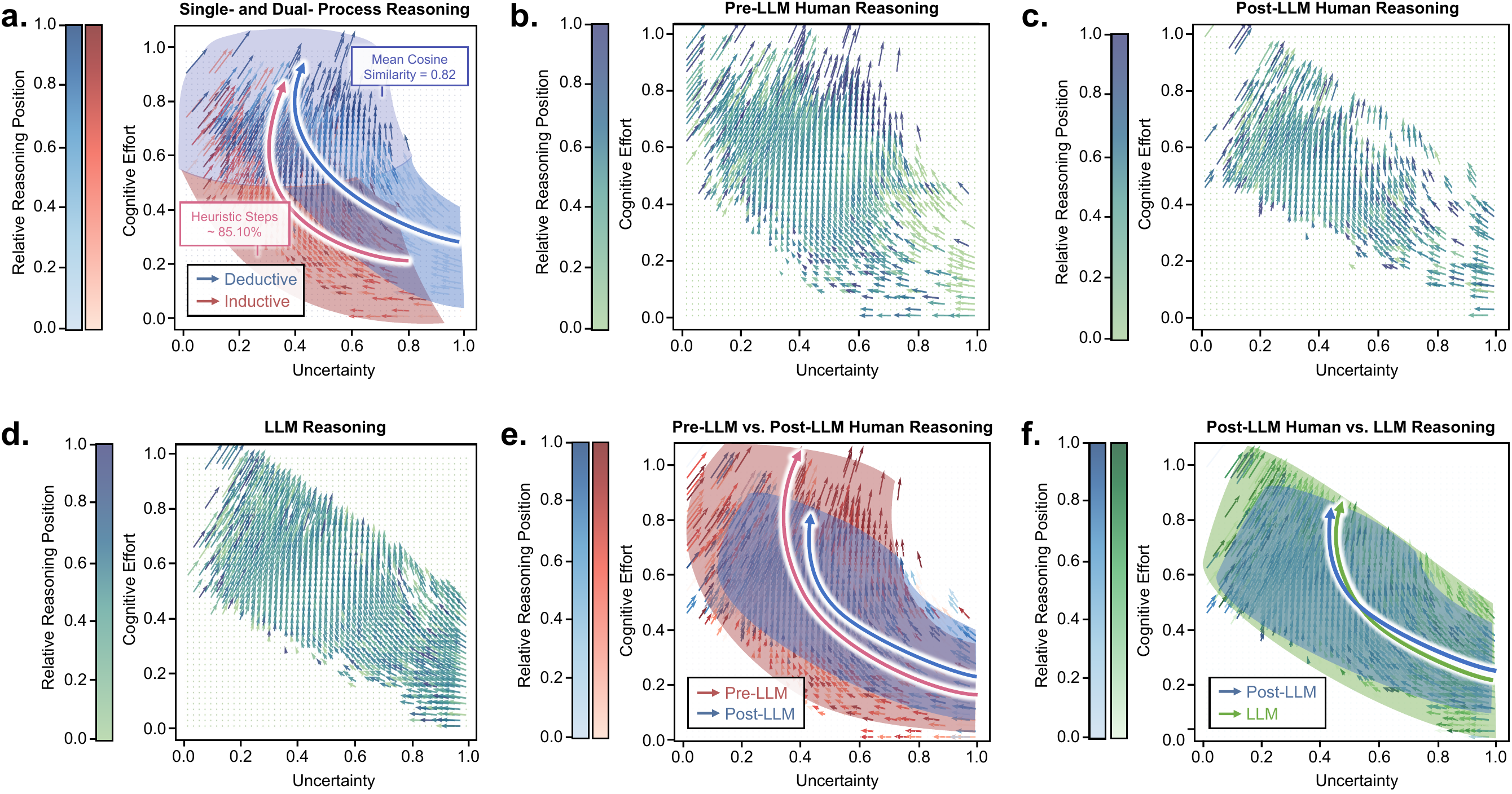}
    \caption{\textbf{Application \modelname{} for Adavanced Psychological Theory Discussion.}\\
    \textit{
    \textbf{a.} Comparison between single- and dual-process theories of reasoning (\(n=1632\)). Dual-process theories posit interacting intuitive and analytic systems, whereas single-process accounts treat reasoning as a graded, continuously integrated computation.
    \textbf{b.} Pre-LLM human reasoning flow (\(n=1667\)). Before the LLM era, human reasoning often began with low effort under high uncertainty and, via analytic integration, progressed toward higher effort with lower uncertainty.
    \textbf{c.} Model reasoning flow (\(n=1537\)). LLMs exhibit a similar trajectory: uncertainty declines as computational depth increases, mirroring human analytic patterns.
    \textbf{d.} Post-LLM human reasoning flow (\(n=1549\)). After interacting with LLMs, human reasoning shows a compressed trajectory, starting at higher effort and lower uncertainty, converging sooner, and often skipping further exploration.
    \textbf{e.} Comparison of pre- and post-LLM human reasoning. Pre-LLM reasoning featured extended exploration and late convergence, whereas post-LLM reasoning stabilizes earlier and is more efficient, signaling a shift from discovery-oriented to synthesis-oriented cognition.
    \textbf{f.} Comparison between post-LLM human and LLM reasoning. Both trajectories now begin at similar levels of uncertainty and effort, suggesting an emerging convergence in reasoning structure across humans and models.
    }
    \vspace{-2mm}}
    \label{fig:human_pre_post}
\end{figure*}

The long-standing debate over \textbf{single- vs. dual-process theories} of reasoning has focused on whether deductive and inductive reasoning stem from distinct systems or a single, continuous mechanism~\citep{evans2008dual,stanovich1999individual}.
Previous studies are inconclusive, some suggest separate neural activations, while others point to their integration and smooth transitions across reasoning stages~\citep{goel2005roles,de2011two,evans2010thinking}.
Here, we argue that this apparent dichotomy can be reconciled through a unified account of reasoning dynamics offered by \modelname{}. 
As shown in Fig.~\ref{fig:human_pre_post}\textit{a}, \modelname{} positions intuitive and analytic modes within a single information-flow continuum.\vspace{-8pt}

\paragraph{Locally, reasoning exhibits dual-process dynamics.}  
Inductive trajectories originate in high-Uncertainty, low-Effort regions (consistent with heuristic exploration) and evolve toward low-Uncertainty, high-Effort states (associated with analytic integration).
Empirically, within the low-Effort regions identified by \modelname{}, \textbf{85.10\% of reasoning steps} can be manually classified as heuristic, indicating a dominant intuitive phase during early reasoning.
This intra-episode shift from intuitive to deliberate processing quantitatively captures dual-process phenomena within individual reasoning paths~\citep{evans2011dual,evans2013dual,moskovitz2022unified}.\vspace{-8pt}

\paragraph{Globally, reasoning follows a single-process flow.}  
Aggregated across tasks and participants, reasoning trajectories exhibit a consistent, monotonic decrease in Uncertainty and a steady increase in Cognitive Effort.
When comparing later-stage reasoning across deductive and inductive datasets, the \textbf{mean cosine similarity between their trajectory vectors reached 0.82}, suggesting strong alignment and structural consistency in the global reasoning flow.
This large-scale regularity supports a single-process framework, indicating that heuristic and analytic modes are dynamically coupled components of a unified system~\citep{de_neys2021on,moskovitz2022unified}.
Thus, \modelname{} shows that dual-process effects arise as local transitions within a global single-process architecture.\vspace{-8pt}

\subsubsection{Human Reasoning Reshaping in the Era of LLMs}

With the rapid development of LLMs, they are increasingly used as essential tools in daily life and work, supporting reasoning and decision-making. Specifically, we utilize GPT-4o~\citep{achiam2023gpt} as a strong model provided stable reasoning.
Concurrent studies indicate that frequent reliance on AI tools can reshape human behavior distributions~\cite{soc15010006,rahwan2019machine,bergman2023chatgpt,long2024ai,shane2024cognitive,shneiderman2020human,bender2020climbing,ChengZhang2025}.
This raises a key question: \textbf{To what extent does reliance on GPT-4o for reasoning lead users to implicitly adopt the model’s reasoning patterns, thereby altering their subsequent reasoning in the model’s absence?}
The resulting patterns are depicted in Fig.~\ref{fig:human_pre_post}\textit{b–f}.\vspace{-8pt}

\paragraph{LLMs are reshaping human reasoning.}  
As shown in Fig.~\ref{fig:human_pre_post}\textit{b–e}, pre-LLM reasoning typically begins with low cognitive effort, reflecting tentative initial intuitions, and increases through exploration and iterative refinement, yielding a low-start, high-end trajectory. In contrast, with extensive reliance on LLMs, reasoning often starts at a higher level of cognitive effort but tends to end lower, producing a high-start, lower-end trajectory. Intuitively, the former builds effort through open-ended search, whereas the latter proceeds along a more constrained pathway that dampens later-stage exploratory effort. Together, these patterns indicate a redistribution of cognitive effort across stages of the reasoning process in the LLM era.\vspace{-8pt}

\paragraph{Post-LLM human reasoning flows closely align with those of LLMs.}  
As shown in Fig.~\ref{fig:human_pre_post}\textit{f}, trajectories produced by GPT-4o largely overlap with human reasoning flows after its release.
This overlap suggests that frequent LLM use not only changes the context in which people reason but also subconsciously encourages users to mimic and internalize model-specific heuristics, promoting convergence between human and machine reasoning.
Post-LLM human trajectories exhibit high initial Cognitive Effort followed by low terminal effort, narrowing the accessible region of the reasoning phase space. Individuals, aligned with LLMs, appear less inclined toward prolonged exploration, and the terminal segment of information flow loses the previously observed exploratory region characterized by reduced uncertainty and elevated Cognitive Effort.

\vspace{-4pt}\section{Conclusion \& Discussion}
In summary, we present a unified, stepwise framework that quantitatively captures the dynamics of human reasoning by tracing information entropy and gain through inferential trajectories. Our approach reconciles classical and probabilistic theories, formalizes reasoning processes in measurable terms, and uncovers individual and group-level cognitive signatures. By applying these tools to discussions on single- versus dual-process theories and comparing human with large language model reasoning, we provide new views for aligning AI with human thought and quantify how LLMs reshape human reasoning.

Future work could extend this framework to real-time neural recordings and dynamic decision-making contexts to further elucidate the neurocognitive mechanisms underlying reasoning. Moreover, \modelname{} may enable the application of this approach in adaptive cognitive training paradigms, allowing for the assessment and enhancement of reasoning skills in educational and clinical settings.
\vspace{-4pt}\section{Method}

This section presents the methodological framework and implementation of our study, which quantitatively models human and model reasoning trajectories within the information phase space defined by \textbf{uncertainty} and \textbf{cognitive effort}.

\vspace{-4pt}\subsection{Detailed \modelname{} Framework}
\label{sec:model}
In this section, we elaborate on the \modelname{} framework introduced in Sec.~\ref{sec:theory}, detailing how it quantitatively computes uncertainty and cognitive effort, and how it formalizes reasoning dynamics as a Hamiltonian system within the information phase space.
\vspace{-4pt}\subsubsection{Quantifying Uncertainty and Cognitive Effort} 
\label{sec:uncertainty_effort}

We next describe in detail how \modelname{} quantitatively computes \textbf{uncertainty} and \textbf{cognitive effort} for each reasoning trajectory. 
These two quantities form the orthogonal dimensions of the information phase space, representing respectively the ambiguity of reasoning states and the cognitive adjustment required between consecutive steps.\vspace{-8pt}

\paragraph{Uncertainty.}  
Given a reasoning step containing $n$ tokens with probabilities $\{p_i\}_{i=1}^{n}$, the uncertainty of that step is defined as the average token-level Shannon entropy~\citep{farquhar2024detecting}:
\begin{equation}
u_t = -\frac{1}{n_t}\sum_{i=1}^{n_t} p_{t,i} \log p_{t,i},
\end{equation}
which reflects the model’s internal uncertainty when generating the $t$-th reasoning step.\vspace{-8pt}

\paragraph{Cognitive Effort.}  
Cognitive effort is defined as the temporal derivative of uncertainty along the reasoning trajectory, representing the rate of entropy change between adjacent steps.  
Expanding $u_t$ and $u_{t-1}$ from the above definition yields:
\begin{equation}
e_t = u_t - u_{t-1} = -\frac{1}{n_t}\sum_{i=1}^{n_t} p_{t,i} \log p_{t,i}
       + \frac{1}{n_{t-1}}\sum_{j=1}^{n_{t-1}} p_{t-1,j} \log p_{t-1,j}.
\label{eq:cognitive_effort}
\end{equation}
This formulation expresses cognitive effort as a difference in token-level entropy expectations between consecutive steps,   
that is, a reorganization of probability mass $\{p_{t,i}\}$ along the reasoning sequence.  
It directly quantifies the magnitude of information restructuring required for cognitive progression,  
aligning with the concept of cognitive effort in cognitive science.\vspace{-8pt}

\paragraph{Normalization.}  
To ensure comparability across reasoning trajectories, we employ two complementary normalization strategies.  
\begin{enumerate}[leftmargin=2ex,topsep=0pt]
  \item \emph{Global normalization}: Both uncertainty and cognitive effort are individually normalized across the entire dataset to the range $[0, 1]$, facilitating comparisons across different datasets or reasoning paradigms by aligning all measurements onto a shared scale.
  \item \emph{Local normalization}: For visualization and intra-sample comparison, each trajectory’s step indices are linearly normalized to $[0, 1]$, preserving its internal temporal dynamics and enabling meaningful comparisons between reasoning steps within the same trajectory.  
\end{enumerate}

Since homeomorphic transformations do not alter the topological structure of trajectories, these normalizations unify the measurement scales without distorting the underlying flow geometry.  
Such rescaling helps us observe the intrinsic dynamical patterns of reasoning trajectories while minimizing the influence of scale differences across datasets or individuals.

\vspace{-4pt}\subsubsection{Liouville Conservation in the Information Phase Space}
\label{sec:liouville_proof}

\paragraph{Notation \& Assumptions:}
Let the reasoning process evolve continuously over the normalized reasoning step 
\(\tau \in [0,1]\). 
At each time \(\tau\), the reasoning state is denoted as:
\begin{equation}
\mathbf{X}_\tau = (u_\tau, e_\tau),
\end{equation}
where \(u_\tau\) represents \textit{uncertainty} (information entropy) and 
\(e_\tau\) represents \textit{cognitive effort} (information gain between consecutive steps).  
The reasoning dynamics is modeled as a continuous flow in the 2D information flow fields (\(\Omega \subset \mathbb{R}^2\)):
\begin{equation}
\dot{\mathbf{X}}_\tau = \mathbf{V}(u_\tau, e_\tau)
= \big(V_1(u_\tau, e_\tau),\, V_2(u_\tau, e_\tau)\big),
\label{eq:flow_field}
\end{equation}
where \(V_1\) and \(V_2\) denote the instantaneous value 
of uncertainty and cognitive effort for human reasoning, respectively. That is, \(V_1 = \dot{u}_\tau\) and \(V_2 = \dot{e}_\tau\).  

Let \(\rho(u_\tau, e_\tau, \tau)\) denote the probability density 
of reasoning states in the information flow fields. Assuming that 
\(\rho\) are change smoothly, it formally has the following assumption:

\textbf{Assumption:}  
Under the \textit{quasi-stationary} condition 
\((\partial_\tau \rho \approx 0)\) and assuming that 
\(\rho\) varies slowly over \(\tau\),  
The density can be treated as locally time-invariant.  
This implies that the evolution of reasoning 
preserves local information volume in expectation.\vspace{-8pt}

\paragraph{Continuity equation and Liouville condition.}
Conservation of probability mass in phase space yields the continuity equation:
\begin{equation}
\frac{\partial \rho}{\partial \tau} 
+ \nabla \!\cdot (\rho \mathbf{V}) = 0,
\quad \text{where }
\nabla \!\cdot (\rho \mathbf{V})
= \frac{\partial (\rho V_1)}{\partial u_\tau}
+ \frac{\partial (\rho V_2)}{\partial e_\tau}.
\label{eq:continuity}
\end{equation}
Under the above assumption, 
\(\partial_\tau \rho \!\approx\! 0\) and 
\(\nabla \rho \!\approx\! 0\),
which simplifies Eq.~\eqref{eq:continuity} to
\begin{equation}
\nabla \!\cdot \mathbf{V}
= \frac{\partial V_1}{\partial u_\tau} 
+ \frac{\partial V_2}{\partial e_\tau} = 0,
\label{eq:liouville_condition}
\end{equation}
known as the \textbf{Liouville condition} for incompressible information flow.
\vspace{-8pt}

\paragraph{Measure preservation and Liouville’s theorem.}
Let \(\Phi_\tau:\Omega \to \Omega\) be the flow generated by a vector field \(\mathbf{V}\), following Eq.~\eqref{eq:flow_field}. This function maps an initial point \(X_0\) to its new position \(\Phi_\tau(X_0)= (f_1, f_2, \dots, f_n)\) after evolving for a time \(\tau\). The derivative of this map is the Jacobian matrix, $J_\tau$:
\begin{equation}
  J_\tau = D\Phi_\tau = 
    \begin{pmatrix}
    \frac{\partial f_1}{\partial x_1} & \frac{\partial f_1}{\partial x_2} & \cdots & \frac{\partial f_1}{\partial x_n} \\
    \frac{\partial f_2}{\partial x_1} & \frac{\partial f_2}{\partial x_2} & \cdots & \frac{\partial f_2}{\partial x_n} \\
    \vdots & \vdots & \ddots & \vdots \\
    \frac{\partial f_n}{\partial x_1} & \frac{\partial f_n}{\partial x_2} & \cdots & \frac{\partial f_n}{\partial x_n}
    \end{pmatrix},
\end{equation} 
which describes how the flow \(\Phi_\tau\) locally applies a linear transformation to the space, such as stretching, compression, or rotation.
Its determinant, \(\det J_\tau\), measures the factor by which the volume of an infinitesimal element changes after evolving for time \(\tau\). The Liouville identity gives the time evolution of this volume change:
\begin{equation}
\frac{d}{d\tau}\!\log \det J_\tau
= (\nabla\!\cdot\!\mathbf{V})(\Phi_\tau(\mathbf{X}_0)) \Rightarrow
\det J_\tau 
= \exp\!\left(\!\int_0^\tau 
\nabla\!\cdot\!\mathbf{V}(\Phi_s(\mathbf{X}_0))\,ds\!\right).
\end{equation}

Hence, $\det J_\tau = 1$ if and only if $\nabla\!\cdot\!\mathbf{V}=0$,  
showing that Eq.~\eqref{eq:liouville_condition} 
is equivalent to phase-space measure preservation, indicating that the information flow is volume-preserving.

\vspace{-4pt}\subsubsection{Discretized Information Phase Space}
\paragraph{Hamiltonian representation of divergence-free flows.}
In any simply connected two-dimensional domain, 
a continuously differentiable divergence-free vector field 
admits a scalar potential \(H(u_\tau, e_\tau)\) such that:
\begin{equation}
\mathbf{V}=\nabla^\perp H
=\left(\frac{\partial H}{\partial e_\tau},
      -\frac{\partial H}{\partial u_\tau}\right),
\label{eq:hamiltonian_field}
\end{equation}
leading to the canonical Hamiltonian system:
\begin{equation}
\dot{u}_\tau = \frac{\partial H}{\partial e_\tau},
\qquad
\dot{e}_\tau = -\,\frac{\partial H}{\partial u_\tau},
\label{eq:canonical_hamiltonian}
\end{equation}
where \(H(u_\tau, e_\tau)\) remains conserved along trajectories,   
that meets:
\begin{equation}
  \frac{dH}{d\tau} = \nabla H \!\cdot\! \dot{\mathbf{X}}_\tau = 0.
\end{equation}
Thus, reasoning dynamics behaves as a Hamiltonian flow 
in  \((u_\tau, e_\tau)\) information flow fields.\vspace{-8pt}

\paragraph{Simplified Hamiltonian Function under the information-theoretic constraint.}
Given the empirical constraint that cognitive effort reflects 
the rate of change of uncertainty (\(e_\tau = \dot{u}_\tau\)),  
we have \(V_1 = e_\tau\).  
Substituting into Eq.~\eqref{eq:liouville_condition} gives 
\(\partial_{e_\tau} V_2 = 0 \Rightarrow V_2 = -U'(u_\tau)\).
Integrating \(\partial_{e_\tau} H = e_\tau\) yields the separable Hamiltonian:
\begin{equation}
H(u_\tau, e_\tau)
=\tfrac{1}{2}e_\tau^2 + U(u_\tau) + C,
\qquad
\dot{u}_\tau = e_\tau, \quad
\dot{e}_\tau = -\,U'(u_\tau),
\label{eq:reduced_hamiltonian}
\end{equation}
for which 
\(\frac{dH}{d\tau} = e_\tau \dot{e}_\tau + U'(u_\tau)\dot{u}_\tau = 0\).  
Hence, \(u_\tau\) and \(e_\tau\) constitute a conjugate pair 
of generalized coordinate and momentum in the information phase space.\vspace{-8pt}

\paragraph{Finite-volume form ensures divergence discretization.}
For a bounded domain 
\(\Omega=[0,1]\!\times\![0,1]\), 
let \(C_{ij}\) be a rectangular control cell centered at 
\((u_i,e_j)\) with sizes \((\Delta u,\Delta e)\).  
By the divergence theorem,
\begin{equation}
(\nabla\!\cdot\!\mathbf{V})_{ij}
=\frac{1}{|C_{ij}|}\!\oint_{\partial C_{ij}}
\mathbf{V}\!\cdot\!\hat{n}\,ds
\approx
\frac{1}{\Delta u}\!
\big[\bar{\dot{u}}_{\,i+\frac12,j}
-\bar{\dot{u}}_{\,i-\frac12,j}\big]
+
\frac{1}{\Delta e}\!
\big[\bar{\dot{e}}_{\,i,j+\frac12}
-\bar{\dot{e}}_{\,i,j-\frac12}\big],
\label{eq:finite_volume}
\end{equation}
where the edge-averaged fluxes are:
\begin{equation}
\bar{\dot{u}}_{\,i\pm\frac12,j}
=\frac{1}{\Delta e}
\!\int_{e_j-\frac{\Delta e}{2}}^{e_j+\frac{\Delta e}{2}}
\dot{u}(u_{i\pm\frac12},e)\,de,
\quad
\bar{\dot{e}}_{\,i,j\pm\frac12}
=\frac{1}{\Delta u}
\!\int_{u_i-\frac{\Delta u}{2}}^{u_i+\frac{\Delta u}{2}}
\dot{e}(u,e_{j\pm\frac12})\,du.
\end{equation}
Approximating edge integrals by the midpoint rule yields (Fig.~\ref{fig:verify}\textit{c} uses this discretization):
\begin{equation}
(\nabla\!\cdot\!\mathbf{V})_{ij}
=\frac{1}{\Delta u}
\big[\dot{u}_{i+\frac12,j}-\dot{u}_{i-\frac12,j}\big]
+\frac{1}{\Delta e}
\big[\dot{e}_{i,j+\frac12}-\dot{e}_{i,j-\frac12}\big]
+O(\Delta u^2+\Delta e^2),
\label{eq:discrete_divergence}
\end{equation}
which provides the standard second-order finite-volume discretization of the Liouville condition in information phase space.

\begin{TakeawayBox}{Summary}
\begin{itemize}[left=2pt,topsep=1pt,itemsep=2pt,parsep=1pt]
  \item Probability conservation $\Rightarrow$ continuity equation (Eq.~\eqref{eq:continuity});
  \item Measure preservation $\Leftrightarrow$ divergence-free condition (Eq.~\eqref{eq:liouville_condition});
  \item In 2D, divergence-free $\Leftrightarrow$ Hamiltonian structure (Eq.~\eqref{eq:hamiltonian_field});
  \item Under $e_\tau=\dot{u}_\tau$, Hamiltonian function $H(u_\tau,e_\tau)=\tfrac{1}{2}e_\tau^2+U(u_\tau)$ (Eq.~\eqref{eq:reduced_hamiltonian});
  \item Finite-volume form (Eqs.~\eqref{eq:finite_volume}--\eqref{eq:discrete_divergence}) ensures divergence discretization.
\end{itemize}

\vspace{2pt}
\end{TakeawayBox}

\subsection{Experimental Setting}
\subsubsection{Experimental Setting under \modelname{}.}
All experiments conducted within our framework follow a unified modelling and data processing configuration.  
Each reasoning trajectory for \modelname{} is encoded using the \textbf{Llama3-8B-Instruct} model, which transforms both the input problem and its step-by-step reasoning process into high-dimensional semantic representations for subsequent computation of information uncertainty and cognitive effort.
Consistent normalization and feature extraction procedures are applied across all reasoning types to ensure comparability.

Unless otherwise specified, all experiments are performed under the same hardware environment and random seed settings to guarantee stability and reproducibility. 
\vspace{-4pt}\subsubsection{Embedding-based Visualization}
\label{sec:method:embedding_tsne}

To examine the geometric structure of reasoning in the information phase space, we compute stepwise embeddings for each trajectory using the ``[CLS]'' representation in BERT~\citep{devlin2019bertpretrainingdeepbidirectional} model.
Each step is encoded as a semantic vector that approximates its latent reasoning state. The resulting collection of embeddings defines a high-dimensional manifold that captures reasoning dynamics.

To visualize the manifold, we apply t-SNE~\citep{maaten2008visualizing} to project the embeddings into two dimensions (\texttt{random\_seed=42}). The resulting map preserves local continuity across successive reasoning steps and reveals global clusters of reasoning patterns. A color gradient encodes each step’s position within its sequence, providing an interpretable view of how model uncertainty and a proxy for cognitive effort vary along the reasoning trajectory.

\vspace{-4pt}\subsubsection{Landscape of Thought for Open-Ended Reasoning}
\label{sec:method:landscape}

To extend the original \textit{Landscape of Thought}~\citep{zhou2025landscape} framework, originally designed only for multiple-choice reasoning, to support open-ended reasoning tasks, we adapted the method through a unified sampling and transformation procedure.
For each open-ended question, we sampled multiple human- or model-generated responses and constructed a \textbf{pseudo multiple-choice set}, where multiple sampled answers as choice set. This design enables consistent embedding and visualization of diverse reasoning trajectories within the same representational space.
For re-implementation, specifically, each reasoning step within these sampled responses was embedded into a semantic vector using BERT. The resulting high-dimensional features were normalized and projected into a two-dimensional manifold using t-SNE to preserve both local continuity and global relational structure across steps.

We then applied a kernel density estimation to capture the overall distribution of reasoning trajectories, generating a continuous “cognitive landscape” that reflects areas of high reasoning convergence and exploratory dispersion. Representative trajectories were visualized by tracing their progression across the landscape, where transparency and color gradients encode step progression from early heuristic exploration to later analytic consolidation.

\begin{figure*}[t]
  \centering
  \includegraphics[width=\linewidth]{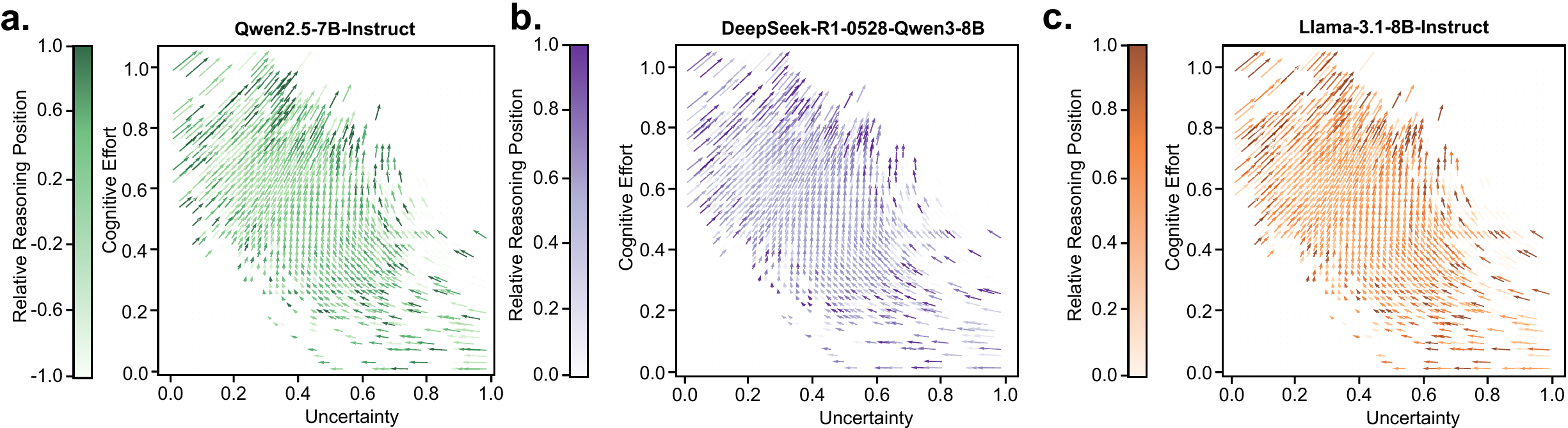}
  \caption{The generalization anlaysis of \modelname{} on Qwen2.5-7B-Instruct (a), DeepSeek-R1-0528-Qwen3-8B (b), Llama-3.1-8B-Instruct (c).\vspace{-8pt}}
  \label{fig:general_flow}
\end{figure*}

\vspace{-4pt}
\subsubsection{\modelname{} on Non-Reasoning Scenarios}
To validate the specificity of the \modelname{} framework to reasoning processes, control experiments were conducted on non-reasoning tasks sconversational dialogue tasks, text summarization, and machine translation. 
For these comparisons, the \textit{Persona-Chat}~\citep{jandaghi2023faithful} dataset was adopted to represent conversational and narrative text generation without explicit reasoning chains, allowing us to evaluate whether the proposed entropy–effort dynamics emerge only in genuine reasoning processes rather than general language modeling behaviors.

Using the same model and data processing pipeline as in the reasoning experiments, we computed uncertainty and cognitive effort for these tasks.  
We then analyzed the resulting trajectories in the information phase space to assess whether they exhibit the same Hamiltonian structure and conservation properties as observed in reasoning tasks.  
This comparison helps to elucidate whether the uncertainty–effort dynamics are unique to reasoning or represent a more general cognitive phenomenon.

\vspace{-4pt}\subsubsection{Generalization Analysis on Different LLMs}
To assess the generalizability of the \modelname{} framework across different LLMs, we replicated our experiments using multiple LLM architectures, including Qwen2.5-7B-Instruct~\citep{qwen2.5}, DeepSeek-R1-0528-Qwen3-8B~\citep{deepseekai2025deepseekr1incentivizingreasoningcapability}, and Llama-3.1-8B-Instruct~\citep{grattafiori2024llama3herdmodels}.  
For each model, we followed the same data processing and analysis procedures as outlined in previous sections.
We computed uncertainty and cognitive effort for reasoning trajectories generated by each model and examined their dynamics in the information phase space.

By comparing the results across different LLMs, we aimed to determine whether the flow structure and properties identified by \modelname{} are consistent features of reasoning processes across diverse model architectures.
As shown in Figure~\ref{fig:general_flow}, the results demonstrate that all tested LLMs exhibit the absolutely same information change dynamics in their reasoning trajectories, supporting the significant robustness and universality of the \modelname{} framework.

\vspace{-4pt}\subsection{Data Collection}
\subsubsection{The Collection of Comprehensive Human Reasoning Data}

To comprehensively validate the generalizability of \modelname{}, we construct an integrated reasoning dataset covering diverse domains and reasoning types.  
As illustrated in Table~\ref{tab:datasets_overview}, this dataset includes more than 100,000 reasoning samples collected from a wide range of existing datasets, spanning mathematics, science, commonsense, logic, and examination-style reasoning.
Each dataset contributes a distinct perspective on reasoning dynamics, enabling us to assess whether the uncertainty–effort framework holds consistently across different cognitive tasks.

\begin{table*}[t]
\centering
\resizebox{\textwidth}{!}{
  \begin{tabular}{lccc}
  \toprule
  \textbf{Dataset} & \textbf{Domain} & \textbf{Reasoning Type} & \textbf{Data Size} \\
  \midrule
  AIME2024~\citep{aime2024} & Mathematics & Deductive & 30 \\
  GSM8K~\citep{gsm8k} & Mathematics & Deductive & 8K \\
  BigGSM~\citep{chen2024unlocking,biggsm}  & Mathematics & Deductive & 610 \\
  MATH~\citep{hendrycksmath2021} & Mathematics & Deductive & 12K \\
  NuminaMathCoT~\citep{numina_math_7b} & Mathematics & Deductive & 16K \\
  OlympiadBench~\citep{he2024olympiadbenchchallengingbenchmarkpromoting} & Mathematics / Science & Deductive \& Inductive & 9K \\
  \midrule
  SciFact~\citep{wadden-etal-2020-fact} & Science & Abductive & 1.4K \\
  PHYBench~\citep{qiu2025phybenchholisticevaluationphysical} & Physics & Inductive \& Abductive & 1K \\
  WorldTree V2~\citep{xie-etal-2020-worldtree} & Science & Deductive \& Inductive & 4.4K \\
  \midrule
  CommonSenseQA~\citep{talmor2019commonsenseqaquestionansweringchallenge} & Commonsense & Inductive \& Abductive & 9K \\
  OpenBookQA~\citep{OpenBookQA2018} & Commonsense / Science & Deductive \& Inductive & 5K \\
  AUQA~\citep{garcia2020datasetbaselinesvisualquestion} & Multimodal / Art & Inductive \& Abductive & 3K \\
  \midrule
  LogiQA~\citep{liu2020logiqachallengedatasetmachine} & Logic & Deductive & 1.8K \\
  CRT-QA~\citep{crtqa2023} & Critical Reasoning & Deductive \& Inductive & 728 \\
  LSAT (AGI-Eval~\citep{zhong2023agieval}) & Logic / Exam & Deductive & 2K \\
  \midrule
  Gaokao (AGI-Eval~\citep{zhong2023agieval}) & Examination & Deductive \& Inductive & 4K \\
  JEC-QA (AGI-Eval~\citep{zhong2023agieval}) & Law / Examination & Deductive \& Inductive & 2K \\
  EKAR~\citep{chen-etal-2022-e} & Analogy & Inductive & 1.1K \\
  GPQA~\citep{rein2023gpqa} & Graduate-level / Knowledge & Deductive \& Abductive & 209 \\
  \midrule
  PRM800K~\citep{lightman2023lets} & Process Supervision (multi-domain) & Deductive \& Abductive & 10K \\
  \bottomrule
  \end{tabular}
}
\caption{
Overview of datasets used for comprehensive reasoning evaluation. 
Each dataset is categorized by its domain, reasoning type, and approximate data size (total $\sim$112K samples).\vspace{-8pt}
}
\label{tab:datasets_overview}
\end{table*}

These datasets collectively span mathematics, science, commonsense, logic, and human-level reasoning tasks, 
providing a comprehensive foundation for analyzing reasoning dynamics across domains.  
All datasets are unified into a consistent JSONL format, where each entry contains a question, a multi-step reasoning process, and a final answer.  
This unified structure allows us to extract token-level entropy for each reasoning step and compute the corresponding cognitive effort as defined in Section~\ref{sec:uncertainty_effort}. 

\vspace{-4pt}\subsubsection{The Collection of Human Reasoning Data with Individual Features}
To better understanding of modelling capabilities on individual features , we present the design and implementation of a large-scale study of human reasoning that captures detailed reasoning trajectories by free-text input alongside individual cognitive characteristics. Specifically, the study comprises two components: participant recruitment and questionnaire design.\vspace{-8pt}

\begin{figure*}[t]
  \centering
  \adjustbox{max width=\linewidth}{
  \includegraphics{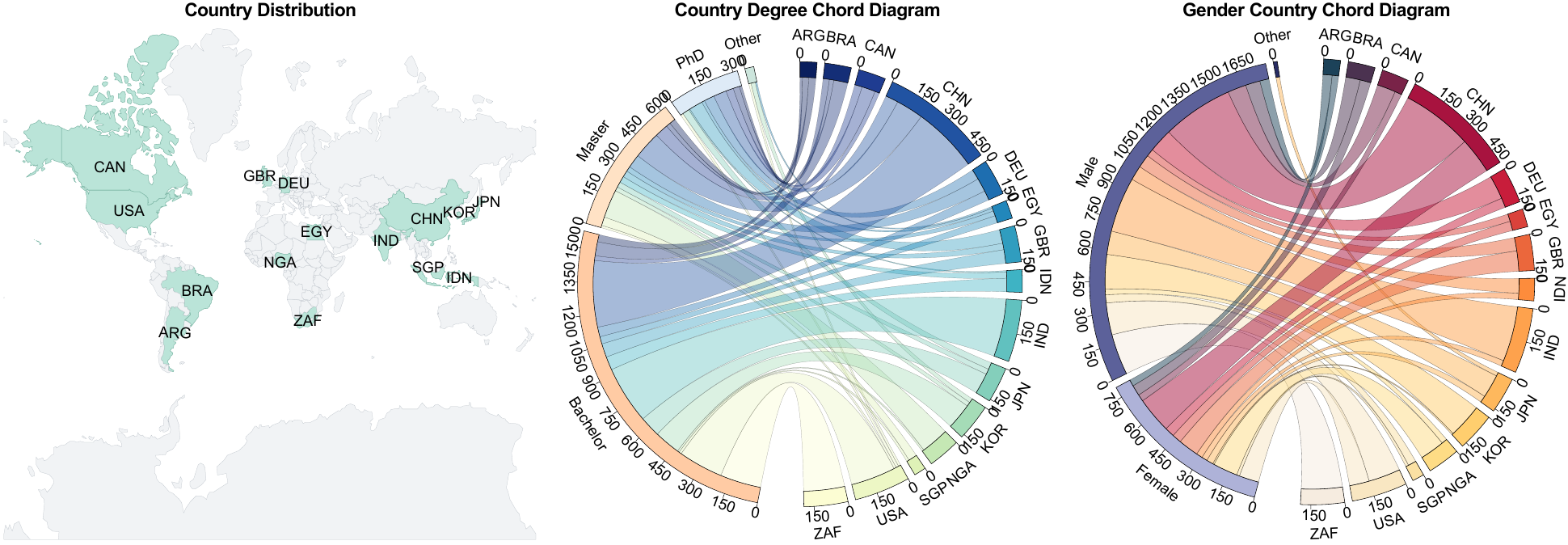}}
  \caption{Geographical and demographic distribution of study participants, illustrating the distribution between nationality and education, and between nationality and gender.}
  \label{fig:country_distribution}
\end{figure*}

\paragraph{Participant Design.}  
We collected 6,452 reasoning trajectories by entrusting commercial companies from participants across 15 countries, 
with the geographical distribution shown in Figure~\ref{fig:country_distribution}\textit{a}. 
The participant pool covered a wide range of educational backgrounds, 
from undergraduate to doctoral level. 
Nationality–education and nationality–gender distributions are visualized 
using chord diagrams in Figure~\ref{fig:country_distribution} \textit{b,c}, 
demonstrating the diversity of population.\vspace{-8pt}

\paragraph{Questionnaire Design.}  
The major problems in the questionnaire are constructed based on the AGI-Eval~\citep{zhong2023agieval} benchmark, 
covering multiple domains including mathematics, medicine, computer science, humanities, and history. 
To ensure balanced coverage of reasoning paradigms, 
we included three reasoning types: deduction, induction, and abduction. 
For each type, six representative problems were randomly sampled from 6K independently for each participant, 
and their order was randomized to control for sequence effects. 
In addition to task performance, we also collected personality information 
based on the Big Five dimensions, using the Ten-Item Personality Inventory (TIPI)~\citep{gosling2003very}, 
a concise 10-item measure that captures the Big Five traits (Extraversion, Agreeableness, Conscientiousness, Emotional Stability, and Openness). 
This enables us to analyze how psychological traits relate to reasoning dynamics.

\vspace{-4pt}\subsubsection{The Collection of Pre-LLM and Post-LLM Human Reasoning Data}
To examine the impact of LLMs on human reasoning, we implemented a two-phase data collection protocol: pre-LLM and post-LLM reasoning tasks.\vspace{-8pt}

\paragraph{Pre-LLM Human Reasoning Data}
In the pre-LLM phase, participants tackled a series of reasoning problems across diverse domains, including mathematics, science, and commonsense reasoning (drawn from the AGI-Eval benchmark). They solved these problems independently and recorded their step-by-step reasoning. To select problems predate the release of GPT-4o, we choose AGI-Eval to avoid data leackage. Moreover, we selected participants with no prior LLM experience.
\vspace{-8pt}

\paragraph{LLM Reasoning Data}
In the LLM phase, we prompted the LLMs used in our study (e.g., GPT-4o~\citep{achiam2023gpt}) with the same AGI-Eval questions. Following \citet{kojima2022large}, each prompt explicitly requested step-by-step reasoning (e.g., ``Let's think step-by-step!''). To capture both typical and diverse reasoning behaviors, we used standard decoding settings (top-p=0.95, temperature=0.6). Following \citet{golovneva2023roscoe}, all model outputs were automatically segmented into reasoning steps and aligned with human step boundaries when available.
\vspace{-8pt}

\paragraph{Post-LLM Human Reasoning Data}
In the post-LLM phase, pre-LLM phase participants first received regular exposure to GPT-4o (daily usage) via guided practice sessions. We then recruited the same cohort or a demographically matched group to revisit a subset of the original problems with similar difficulty and categories, avoiding exact duplicates to prevent knowledge leakage. Participants are also banned  from the use of LLMs while documenting their step-by-step processes to show the change of the human reasoning process.

This two-phase design, applied to the same problem subset and comparable participant demographics, enabled direct comparison of human reasoning trajectories before and after LLM exposure. It thus provides insights into how LLMs influence human cognitive processes and reasoning patterns.

\vspace{-4pt}\section{Ethical Considerations}
All procedures involving human participants were reviewed, and informed consent was obtained before data collection. All data were anonymized to ensure participant privacy. Participants received fair compensation in accordance with institutional guidelines. A professional labeling company annotated the reasoning data at a rate of \$2.5 per participant, and all labelers held at least a college degree.

\vspace{-4pt}\section{Code Availability}
The code and relevant data are available at \url{https://github.com/LightChen233/Human-Reasoning-Modeling}.

\bibliographystyle{plainnat}
\bibliography{ref}

\end{document}